\documentclass[runningheads]{llncs}
\usepackage{graphicx}
\usepackage{tikz}
\usepackage{comment}
\usepackage{amsmath,amssymb} 
\usepackage{color}

\usepackage[accsupp]{axessibility} 

\newcommand{\etal}{\textit{et al.}}

\begin{document}
\pagestyle{headings}
\mainmatter
\def\ECCVSubNumber{100} 

\title{Bounded Future MS-TCN++ for surgical gesture recognition} 



\titlerunning{Bounded Future MS-TCN++ for surgical gesture recognition}
%
\author{Adam Goldbraikh\inst{1} \and
Netanell Avisdris\inst{2} \and
Carla M. Pugh\inst{3} \and
Shlomi Laufer \inst{4}}
\authorrunning{Adam Goldbraikh et al.}
%
\institute{Applied Mathematics Department, Technion – Israel Institute of Technology, Haifa 3200003, Israel \\ \email{sgoadam@campus.technion.ac.il} \and
School of Computer Science and Engineering, The Hebrew U. of Jerusalem, Israel
\email{netana03@cs.huji.ac.il}
\and
Stanford University, School of Medicine
Stanford, CA, USA \\
\email{cpugh@stanford.edu}
\and Faculty of Industrial Engineering and Management at the Technion – Israel Institute of Technology, Haifa 3200003, Israel\\
\email{laufer@technion.ac.il}
}

\maketitle

\begin{abstract}
In recent times there is a growing development of video-based applications for surgical purposes. Part of these applications can work offline after the end of the procedure, other applications must react immediately. However, there are cases where the response should be done during the procedure but some delay is acceptable. In the literature, the online-offline performance gap is known. Our goal in this study was to learn the performance-delay trade-off and design an MS-TCN++-based algorithm that can utilize this trade-off.
To this aim, we used our open surgery simulation data-set containing 96 videos of 24 participants that perform a suturing task on a variable tissue simulator. In this study, we used video data captured from the side view. The Networks were trained to identify the performed surgical gestures. The naive approach is to reduce the MS-TCN++ depth, as a result, the receptive field is reduced, and also the number of required future frames is also reduced. We showed that this method is sub-optimal, mainly in the small delay cases. The second method was to limit the accessible future in each temporal convolution. This way, we have flexibility in the network design and as a result, we achieve significantly better performance than in the naive approach.

\keywords{Surgical simulation \and Surgical gesture recognition \and Online algorithms}
\end{abstract}

\section{Introduction}

Surgical data science is an emerging scientific area \cite{maier2017surgical,maier2022surgical}. It explores new ways to capture, organize and analyze data with the goal of improving the quality of interventional healthcare. With the increased presence of video in the operating room, there is a growing interest in using computer vision and artificial intelligence (AI) to improve the quality, safety, and efficiency of the modern operating room. 

A common approach for workflow analysis is to use a two-stage system. The first stage is a feature extractor such as I3D \cite{carreira2017quo} or ResNet50 \cite{czempiel2020tecno,ramesh2021multi}. The next stage usually includes temporal filtering. The temporal filtering may include recurrent neural networks such as LSTM \cite{ullah2017action,yue2015beyond,donahue2015long}, and temporal convolutional networks (TCN) such as MS-TCN++ \cite{li2020mstcnpp} or transformers \cite{chinayi_ASformer,neimark2021video}.

Automatic workflow analysis of surgical videos has many potential applications. It may assist in an automatic surgical video summarizing \cite{avellino2021surgical,lux2010novel}, progress monitoring \cite{padoy2019machine}, and the prediction of remaining surgery duration \cite{twinanda2018rsdnet}. The development of robotic scrub nurses also depends on the automatic analysis of surgical video data \cite{jacob2011gesture,sun2021robotic}. In addition, video data is used for the assessment of surgical skills \cite{goldbraikh2022video,basiev2022open,funke2019video} and identifying errors \cite{jung2020first,mascagni2022artificial,mascagni2021computer}.

Traditionally systems are divided into causal and acausal. However, not all applications require this dichotomic strategy. For example, the prediction of the remaining surgery duration may tolerate some delay if this delay ensures a more stable and accurate estimation. On the other hand, a robotic scrub nurse will require real-time information. In general, any acausal system may be transformed into a causal system if a sufficient delay is allowed. Where in the extreme case, the delay would be the entire video. This study aims to find the optimal system, given a constraint on the amount of delay allowed.

Many studies use Multi-Stage Temporal Convolutional Networks (MS-TCN) for workflow analysis \cite{li2020mstcnpp}. It has both causal and acausal implementations. The network's number of layers and structure defines the size of its receptive field. In the causal case, the receptive field depends on past data. On the other hand, in the acausal case, half of the receptive field depends on future data and half on the past. Assume a fixed amount of delay $T$ is allowed. A naive approach would be to use an acausal network with a receptive field $2\cdot T$. However, this limits the number of layers in the network and may provide sub-optimal results. In this study, we develop and assess a network with an asymmetric receptive field. Thus we may develop a network with all the required layers while holding the constraint on the delay time $T$. This network will be called call Bounded Future MS-TCN++ (BF-MS-TCN++). We will compare this method to the naive approach that reduces the receptive field's size by changing the network's depth. The naive approach will be coined Reduced Receptive Field MS-TCN++ (RR-MS-TCN++). We perform gesture recognition using video from an open surgery simulator to evaluate our method.

The main contribution of our work is the development of an MS-TCN++ with a bounded future window, which makes it possible to improve the causal network performance by delaying the return of output at a predetermined time. In addition, we evaluated a causal and acausal video-based MS-TCN++ for gesture recognition on the open surgery suturing simulation data.

\section{Related Work}

Lea \etal \cite{lea2017temporal} was the first to study TCN's ability to identify and segment human actions. Using TCN, they segmented several non-surgical data sets such as 50 Salads, GETA, MERL Shopping, and Georgia Tech Egocentric Activities. They implemented causal and acausal TCNs and compared their performance on the MERL data set. The acausal solution provided superior results. They also outperform a previous study that uses an LSTM as a causal system and a Bidirectional LSTM as an acausal system. In TeCNO \cite{czempiel2020tecno} causal Multi-Stage Temporal Convolutional Networks (MS-TCN) were used for surgical phase recognition. Two data-sets of laparoscopic cholecystectomy were used for evaluation. The MS-TCN outperformed various state-of-the-art LSTM approaches. In another study \cite{ramesh2021multi}, this work was expanded to a multi-task network and was used for step and phase recognition of gastric bypass procedures. 

In one study, the performance of an acausal TCN was assessed. The analysis included both non-surgical action segmentation datasets as well as a dataset of a simulator for robotic surgery\cite{lea2016temporal}. Zhang \etal \cite{zhang2021swnet} used a Convolutional Network to extract local temporal information and an MS-TCN to capture global temporal information. They used acausal implantation to perform Sleeve Gastrectomy surgical workflow recognition. Not all studies use a separate network for capturing temporal information. In Funke \etal \cite{funke2019using} a 3D convolutional neural networks was used. In this study, they used the sliding window approach to evaluate different look-ahead times. 

The use of TCN is not limited to video segmentation. It has been studied in the context of speech analysis as well \cite{pandey2019tcnn}. In this context, the relationship between delay and accuracy has been assessed \cite{peddinti2015time}.

\section{Methods}

\subsection{Dataset}
Eleven medical students, one resident, and thirteen attending surgeons participated in the study. Their task was to place three interrupted instrument-tied sutures on two opposing pieces of the material. Various materials can simulate different types of tissues; for example, in this study, we used tissue paper to simulate a friable tissue and a rubber balloon to simulate an artery. The participants performed the task on each material twice. Thus, the data set contains 100 procedures, each approximately 2-6 minutes long. One surgeon was left-handed and was excluded from this study. Thus, this study includes a total of 96 procedures.
Video data were captured in $30$ frames per second, using two cameras, providing top and side views. In addition to video, kinematic data were collected using electromagnetic motion sensors (Ascension, trakSTAR). For this study, we only use the side-view camera.
We perform a gesture recognition task, identifying the most subtle surgical activities within the surgical activity recognition task family. Six surgical gestures were defined: G0 - nonspecific motion, G1 - Needle passing, G2 - Pull the suture, G3 - Instrumental tie, G4 - Lay the knot G5 - Cut the suture. The video data were labeled using Behavioral Observation Research Interactive Software (BORIS) \cite{boris}.

\subsection{Architecture}
MS-TCN++ \cite{li2020mstcnpp} is a \emph{temporal convolutional network} (TCN) designed for video data activity recognition. The input for this network is a vector of features extracted from the raw video using a CNN, such as I3D \cite{carreira2017quo}. The video length is not predetermined. Let us assume that the video is given in 30 frames per second and contains $T$ frames, namely, T is a parameter of a specific video. In the following sections, we will describe the different components of the MS-TCN++ and the modifications made for the BF-MS-TCN++. It should be noted that the naive approach, RR-MS-TCN++ has the same structure as the acausal MS-TCN++ and was coined with a unique name to emphasize the limitation on the receptive field size.

The architecture of MS-TCN++ is structured from two main modules: the prediction generator and the refinement. For the sake of simplicity, we will first describe the refinement module structure and then the prediction generator.

\subsubsection{Refinement module:}
\begin{figure}
 \centering
  \includegraphics[width=.9\textwidth]{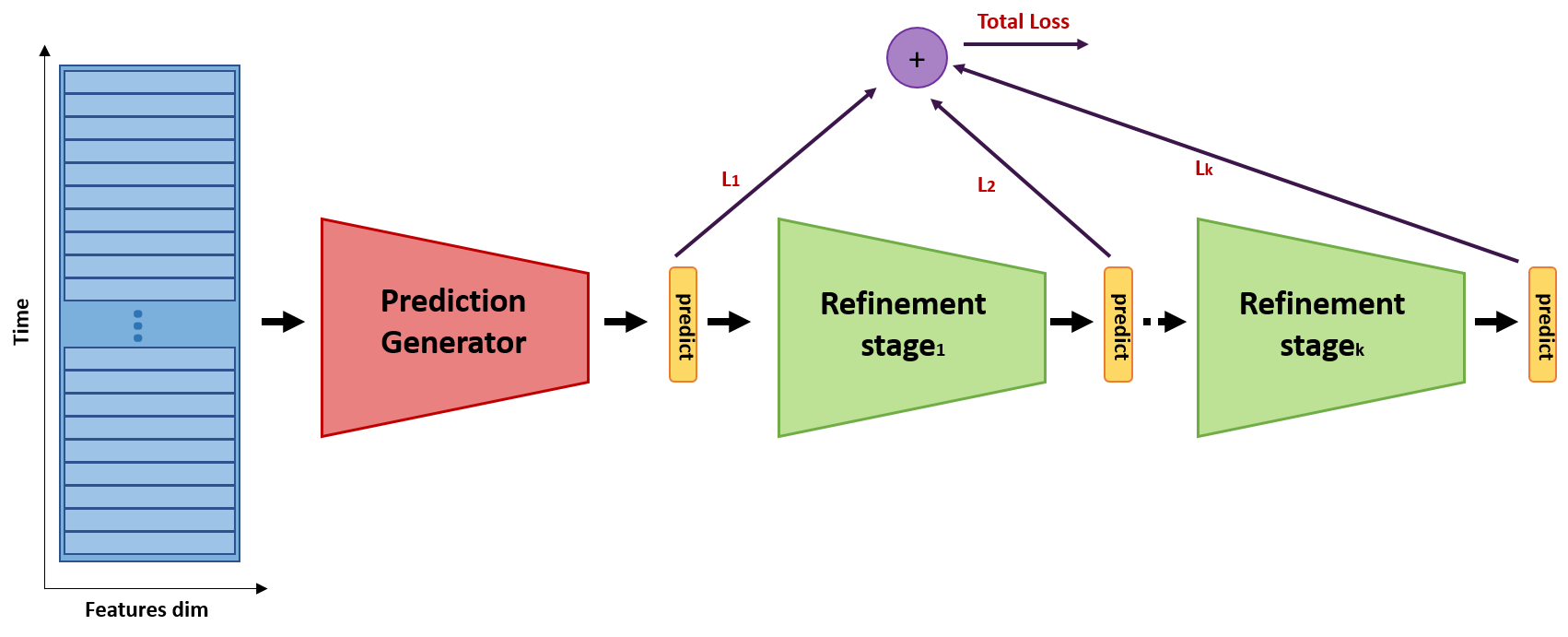}
 \caption{General structure of MS-TCN++. The input is a vector of features (blue). It composes of multiple stages, where each stage predicts the frame segmentation. The first stage is the prediction generator (red) and the other are refinement stages (green), which can be any number ($k>=0$) of them. The loss is computed over all stages' predictions.}
     \label{Figure:MS-TCN++}

 \end{figure}

 As shown in Fig. \ref{Figure:MS-TCN++}, the refinement module includes several refinement stages, where the output of each stage is the input of the next. The refinement stage is a pure TCN that uses \emph{dilated residual layers} (DRL). This allows the module to handle varying input lengths. To match the input dimensions of the stage with the number of feature maps, the input of the refinement stage passes through a $1 \times 1 $ convolutional layer. Then these features are fed into several DLRs where the dilation factor is doubled in each layer. The dilation factor determines the distance between kernel elements, such that a dilation of 1 means that the kernel is dense. 
 Formally, the dilation factor is defined as $ \delta(\ell) = 2^{\ell - 1}: \ell \in \{1,2, \dots L\}$, where $\ell$ is the layer number and $L$ is the total number of layers in the stage. In MS-TCN++, the DRL is constructed from an acausal dilated temporal convolutional layer (DTCL), with a kernel size of 3, followed by ReLU activation and then a $1 \times 1 $ convolutional layer. The input of the block is then added to the result by a standard residual connection, which is the layer's output. To get the prediction probabilities, the output of the last DRL passes through a prediction head which includes a $1 \times 1$ convolution, to adjust the number of channels to the number of classes, and is activated by a softmax.
 
\subsubsection{Prediction Generation module:}
\label{subsubsection:pg}

The prediction generator consists of only one prediction generation stage. The general structure of this stage is similar to the refinement stage; however, instead of a DRLs, it has a \emph{dual dilated residual layers} (DDRLs). Let's consider layer $\ell \in \{1,2, \dots, L\}$. The input of the DDRL is entered into two DTCLs, one with a dilation factor of $\delta_{1}(\ell) = 2^{\ell - 1}$ and the other with a dilation factor of $\delta_{2}(\ell) = 2^{L - \ell}$. Then, the outputs of the two DTCLs are merged by concatenation of the feature in the channel dimension, followed by a 1D convolutional layer to reduce the number of channels back to the constant number of feature maps. This output passes through a ReLU activation and an additional 1D convolutional layer before the residual connection. For a formal definition of MS-TCN++ stages and modules, see \cite{li2020mstcnpp}.

\subsubsection{Future window size analysis}
\label{subsubsection:temporal_convolution_layers}
As MS-TCN++ is a temporal convolutional network with different dilatation among the layers, analyzing the temporal dependence is not trivial. In order to determine the number of future frames involved in the output calculation, careful mathematical analysis is required. Calculating the number of future frames required is equivalent to the output delay of our system.

In the naive approach, RR-MS-TCN++, the number of layers of the network governs the receptive field and the future window. The BF-MS-TCN++ is based on the limitation of the accessible future frames in each temporal convolution; hence the name Bounded Future. This section aims to analyze both methods and calculate desired future window.

The input and the output of DRLs and DDRLs ((D)DRL) have the same dimensions of $N_f \times T$, where $N_f$ represents the number of feature maps in the vector encoding the frame and $T$ is the number of frames in the video. We assume that for every (D)DRL, the vector in the $t$ index represents features that correspond to the frame number $t$. However, in the acausal case, the features of time $t$ can be influenced by a future time point of the previous layer output. 
In MS-TCN++, (D)DRL has symmetrical DTCLs with a kernel size of $3$. The layer's input $\ell$ is padded by $2 \times \delta(\ell) $ zero vectors to ensure the output dimensions are equal to the input dimensions. 
A symmetric convolution is created by padding the input with $\delta(\ell)$ number of zeros vectors before and after it.
The result is that half of each layer's receptive field represents the past and the other half represents the future.
To obtain a causal solution, all $2 \times \delta(\ell) $ zero vectors are added before the input. As a result, the entire receptive field is based on the past time. This method has been used in \cite{lea2017temporal,czempiel2020tecno}.

Let $S= \{PG,R, Total\}$ be a set of symbols, where $PG$ represents relation to the prediction generator, $R$ to the refinement stage, and $Total$ to the entire network.
Let $L_{s} : s\in S$ be the number of (D)DRLs in some stage or in the entire network.
The number of refinement stages in the network is $N_R$. Note that we assume that all refinement stages are identical, and that $L_{Total} = L_{PG} + N_R \cdot L_{R}$.
Let the ordered set $\mathcal{L}_{total} = (1,2, \dots, L_{Total})$, where $\ell \in \mathcal{L}_{total} $ represents the index of $\ell^{th}$ (D)DRL in the order it appears in the network.
Given integer $x$, $[x]$ denotes the set of integers satisfying $ [x] = \{1,\cdots, x\}$.

\begin{figure}[ht]
 \centering
  \includegraphics[width=.6\textwidth]{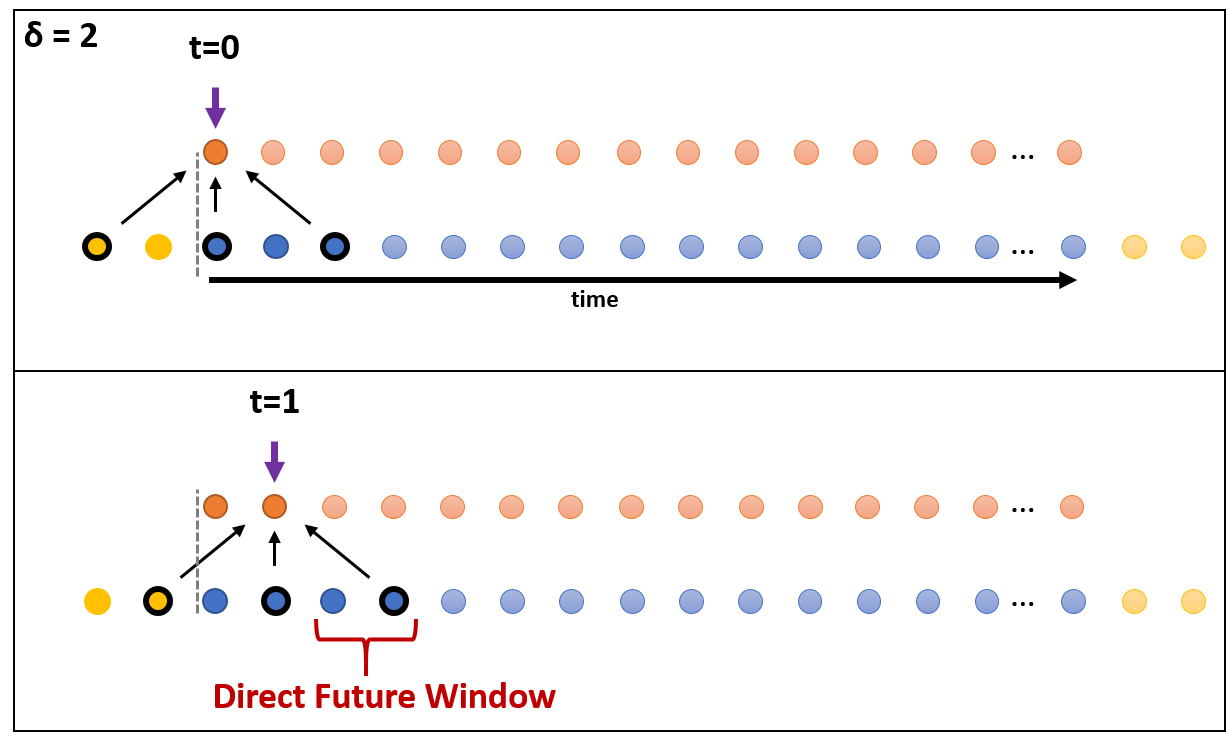}
 \caption{Illustration of Direct Future Window of a dilated temporal convolutional layer (DTCL) with $\delta = 2$, for first (upper part) and second (lower part) timeframes. Blue dots denote input features, yellow dots denote padding needed for DTCL, and orange dots denotes the output of DTCL. }
     \label{Figure:DFW}
 \end{figure}

\begin{definition} Let DTCL $\phi$ with dilation factor of $\delta(\phi)$.
The \emph{Direct Future Window} of a $\phi$ is $m \in [\delta(\phi)]$ if and only if the number of the padding vectors after the layers input is $m$ and number of padding vectors before the vector is $2m - \delta(\phi)$.
The function $DFW(\phi) = m $ gets a DTCL and returns it's direct future window.
\end{definition}
The definition of a Direct Future Window is shown in Fig. \ref{Figure:DFW}.

\subsubsection{Reduced Receptive Field MS-TCN++ case:}
\label{subsubsection:MS-TCN++_case}
In the DRLs, in the refinement stages, the direct future window is equal to the direct future window of it's DTCL. Formally, let $\ell \in [L_{R}]$ a DRL, that contains a DTCL $\phi$. The direct future window of this layer is given by $ \mathcal{DFW}_{R}(\ell) = DFW(\phi) $, where the subscript $R$ indicates an association with a refinement stage. However, each DDRL contains two different DTCLs $\phi_1$ and $\phi_2$, as described in section \ref{subsubsection:pg}. Consider a DDRL $\ell \in [L_{PG}]$. The direct future window of this layers is given by equation \ref{eq:PD_DFW}.
\begin{equation}
\label{eq:PD_DFW}
 \mathcal{DFW}_{PG}(\ell) = \max\{DFW(\phi_1),DFW(\phi_2) \}
\end{equation}

\begin{figure}[ht]
 \centering
  \includegraphics[width=.7\textwidth]{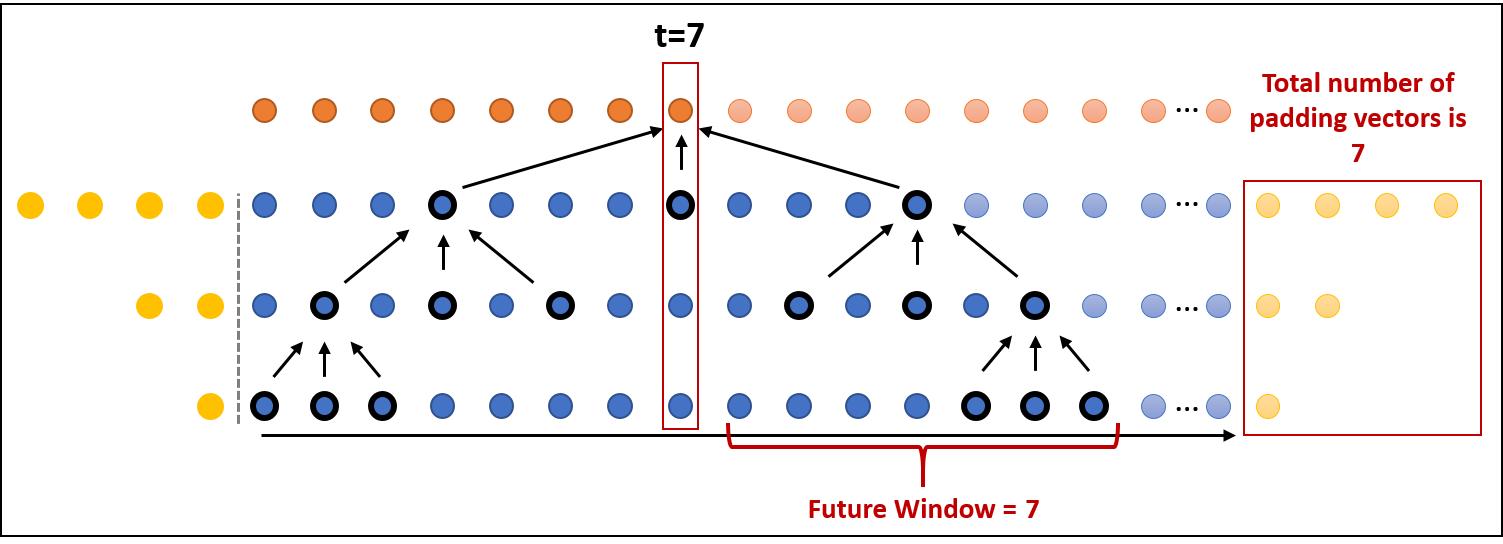}
 \caption{Illustration of Future Window in a refinement stage with three dilated residual layers (DRLs).}
     \label{Figure:FW}
 \end{figure}

\begin{definition}
\label{def:FW}
The \emph{Future Window} of layer $\ell \in \mathcal{L}_{total}$ defined as follows:
\[FW(\ell) = \sum_{i\in [\ell]}\mathcal{DFW}(i)\]
\end{definition}
Fig. \ref{Figure:FW} shows how direct future windows are aggregated to the future window.

Based on definition \ref{def:FW}, the future window of the RR-MS-TCN++ network is defined in equation \ref{eq:FW_MS_TCN++}.

\begin{equation}
\label{eq:FW_MS_TCN++}
FW^{RR}(L_{Total}) = \sum_{\ell \in [ L_{PG}]}\max\{2^{\ell -1}, 2^{L_{PG}-\ell}\} + N_{R} \cdot \sum_{\ell \in [L_{R}]}2^{\ell -1}
\end{equation}

Note that superscript RR indicates that the network is RR-MS-TCN++.
According to definition \ref{def:FW}, equation \ref{eq:FW_MS_TCN++} is obtained by summing the prediction generator and refinement stages separately. In addition, the fact that in the prediction generator, for every DDRL $\ell \in [L_{PG}]$ there exists two DTCLs $\phi_1,\phi_2$ that satisfied $DFW(\phi_1)= 2^{\ell-1}$, and $DFW(\phi_2)=2^{L_{PG}-\ell}$, yields that the direct future window of the DDRL is the maximum between these terms, as defined in equation \ref{eq:DFW_PG_residual}. We take the maximum since the direct future window is determined by the furthest feature in the layer's input that participates in the calculation.

\begin{equation}
\label{eq:DFW_PG_residual}
\mathcal{DFW}_{PG}(\ell)= \max\{2^{\ell -1}, 2^{L_{PG}-\ell}\}
\end{equation}

and for some DRL $\ell \in [L_{R}]$ in the refinement $\mathcal{DFW}(\ell)= 2^{\ell-1}$.

\subsubsection{Bounded Future MS-TCN++ case:}
Let $w_{max}$ be a bounding parameter that bounds the size of the direct future window.
We determine that the direct future window of every DRL $\ell \in [L_R]$, in the refinement stage of the BF-MS-TCN++, is given by equation \ref{eq:DFW_R_BF}.

\begin{equation}
\mathcal{DFW}^{BF}_{R}(\ell) =\min\{ w_{max}, \delta(\ell)\} =\min\{ w_{max}, 2^{\ell -1}\}
\label{eq:DFW_R_BF}
\end{equation}

The superscript $BF$ indicates that the network is BF-MS-TCN++ and the subscript $R$ indicates that this belongs to the refinement stage, where the replacement of $R$ with $PG$ indicates association with the prediction generator.

\begin{figure}[ht]
 \centering
  \includegraphics[width=.9\textwidth]{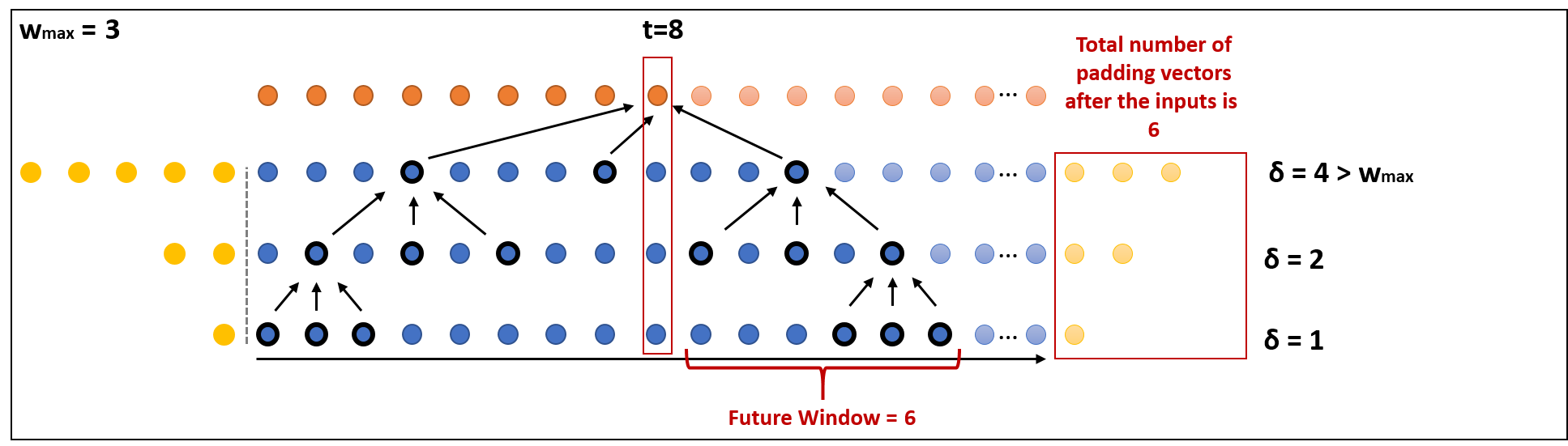}
 \caption{Illustration of a Future Window and asymmetry in padding in a refinement stage of Bounded Future MS-TCN++ with three dilated residual layers (DRLs) and $w_{max} = 3$. In the last layer, since the dilation factor $\delta$ is larger than $w_{max}$, the number of padding vectors (yellow) before and after the sequence is different.}
 
     \label{Figure:bouded_FW}
 \end{figure}

Fig.\ref{Figure:bouded_FW} illustrates how the convolution's symmetry is broken in the case of $\delta(\ell) > w_{max}$.

The future window of the refinement stage is given by equation \ref{eq:FW_R_residual_BF}.
\begin{equation}
\label{eq:FW_R_residual_BF}
FW^{BF}_{R}(L_R)=\sum_{\ell \in [L_R]}\mathcal{DFW}^{BF}_{R}(\ell) = \sum_{\ell \in [L_R]}\min\{ w_{max}, 2^{\ell -1} \}
\end{equation}

For the DDRL $\ell \in [L_{PG}]$ of the prediction generators, the direct future window is given by equation \ref{eq:DFW_PG_BF}

\begin{equation}
\label{eq:DFW_PG_BF}
\mathcal{DFW}^{BF}_{PG}(\ell) = \max\{min\{ w_{max}, 2^{\ell -1} \},min\{ w_{max}, 2^{L_{PG}-\ell} \}\}
\end{equation}

Hence the future window of the prediction generator for a causal network with delay is given by equation \ref{eq:FW_PG_residual_BF}.

\begin{equation}
\label{eq:FW_PG_residual_BF}
FW^{BF}_{PG}(L_{PG}) \sum_{\ell \in [L_{PG}]}\mathcal{DFW}^{BF}_{PG}(\ell) =
\end{equation}

\[ = \sum_{\ell \in [L_{PG}]} \max\{\min\{w_{max},\delta_{1}(\ell)\},\min\{w_{max},\delta_{2}(\ell)\}\}\]

This leads to equation \ref{eq:FW_residual_BF} which presents the future window for the entire network.

\begin{equation}
\label{eq:FW_residual_BF}
FW^{BF}(L_{Total}) = \sum_{\ell \in [ L_{PG}]}\max\{\min\{w_{max},2^{\ell -1}\},\min\{w_{max},2^{L_{PG} - \ell}\}\} +
\end{equation}
\[
 + N_{R} \cdot \sum_{\ell \in [L_{R}]}\min\{{w_{max},2^{\ell -1}}\}
\]

Note that a network with $w_{max} = 0$ is a causal network, which may have a fully online implementation.

\subsection{Feature extractor implementation details}
As a first step, we trained a (2D) Efficientnetv2 Medium \cite{tan2021efficientnetv2} in a frame-wise manner; namely, the label of each frame is its gesture. The input was video frames with a resolution of $224\times224$ pixels. For each epoch, we sampled in a class-balanced manner 2400 frames, such that each gesture appears equally among the sampled frames. The frames were augmented with corner cropping, scale jittering, and random rotation. The network was trained for 100 epochs, with a batch size of 32. Cross-entropy loss was used. 
All the experiments were trained using an Adam optimizer, with an initial learning rate of $0.00025$ that was multiplied by a factor of $0.2$ after 50 epochs, and decay rates of $\beta_1 = 0.9$ and $\beta_2= 0.999$. The code of this part is based on the code provided by Funke \etal \cite{funke2019using}. After the individual training of each split, the one before the last linear layer was extracted and used as a feature map for the MS-TCN++.
Training and evaluation of all networks were done using a DGX cluster with 8 Nvidia A100 GPUs. 

\section{Experimental setup and Results}
\subsection{Experimental setup}
In order to evaluate the effect of the delay on the performance of online algorithms we compare two methods, the naive \textbf{RR-MS-TCN++} and our \textbf{BF-MS-TCN++}.
To this aim, we performed a hyperparameter search. For both networks, the number of refinement stages and the number of (D)DRLs inside the stages affect the total receptive field and hence the future window. The uniqueness of the BF-MS-TCN++ is that the future window can be limited by the bounding parameter $W_{max}$ as well, regardless of the values of the other parameters. In our search, we forced the number of DRLs in the prediction generator to be equal to the number of DRLs in the refinement stages, namely $L_{PG} = L_{R} = L$.

To this end, for the RR-MS-TCN++ network, the number of DRLs $L$ included in the search was in the range of $\{2,3,4,5,6,8,10\}$. The number of refinement stages $N_R$ was in the range of $\{0, 1, 2, 3\}$. For the BF-MS-TCN++, the search included two grids. In the first grid, the values of $w_{max}$ were in the range of $\{0, 1, 2, 3, 6, 7, 8, 10, 12, 13, 14,15, 16, 17, 20 \}$, where $0$ represents a online algorithm. The number of DRLs $L$ was in the range of $\{6,8,10\}$, and the number of refinement stages $N_R$ was in the of $\{0, 1, 2, 3\}$. In the second grid, the values of $w_{max}$ were in the range of $\{1, 3, 7, 10, 12,15, 17\}$. The number of DRLs $L$ was in the range of $\{2,3,4,5\}$, and the number of refinement stages $N_R$ was in the of $\{0, 1, 2, 3\}$. In total, we performed 320 experiments. The rest of the hyperparameters remained constant, where the learning rate was $0.001$, dropout probability was $0.5$, the number of feature maps was $128$, batch size of $2$ videos, the number of epochs was $40$, and the loss function was the standard MS-TCN++ loss with hyperparameters $\tau^{2}=16$ and $\lambda = 1$. All experiments were trained with an Adam optimizer with the default parameters. Training and evaluation were done using a DGX cluster with 8 Nvidia A100 GPUs.

\subsection{Evaluation method}
 We evaluated the models using 5-fold cross-validation. All videos of a specific participant were in the same group (leave-n-users-out approach). The videos assigned to the fold served as the test set of that fold. The remaining participants' videos were divided into train and validation sets.

 The validation set for fold $i \in [5]$ consists of 3 participants from fold $(i+1)\mod5$. 
 Namely, for each fold, 12 videos were used as a validation set (3 participants $\times$ 4 repetitions). With this method, we create unique validation sets for each fold.
 The stopping point during training was determined based on the best F1@50 score on the validation set.
 The metrics were calculated for each video separately, so the reported results for each metric are the mean and the standard deviation across all 96 videos when they were in the test set.
 
\subsection{Evaluation Metrics}
We divide the evaluation metrics into two types: segmentation metrics and frame-wise metrics.
The segmentation metrics contain F1@k where $k\in \{10, 25, 50\}$ \cite{lea2017temporal}, and the edit distance \cite{lea2016learning}. While the frame-wise metrics included Macro-F1 \cite{huang2020sample} and Accuracy. We calculated each metric for each video and then averaged them across all videos.\\[1ex]

\textbf{F1@k:} The intersection over union (IoU) between the predicted segments and the ground truth was calculated for each segment. If there is a ground truth segment with IoU greater than $k$, that ground truth segment is marked as true positive and is not available for future use. Otherwise, the predicted segment has been defined as a false positive. Based on these calculations, the F1 score was determined.
\\

\textbf{Segmental Edit Distance:} The segmental edit distance is based on the Levenshtein distance, where the role of a single character is taken by segments of the activity. The segmental edit distance was calculated for all gesture segments in the video and normalized by dividing it by the maximum between the ground truth and prediction lengths.\\

\textbf{Frame-Wise Metrics: }
We calculated the Accuracy and the multi-class F1-Macro scores as used in \cite{goldbraikh2022sensors}.

\subsection{Experimental studies}
We performed four studies: (1) Baseline estimation; (2) Performance comparison; (3) Network hyperparameter importance; and (4) Competitive analysis.

(1) First, we try to estimate the best fully casual and acausal networks, which will serve as a baseline. 

(2) In the \textit{Performance comparison} study we compare directly between the performance of \textit{Reduced Receptive Field MS-TCN++} and our \textit{Bounded Future MS-TCN++}, with respect to Future Window size. We defined 12 Future Window intervals: $[0, 0.001, 0.125, 0.25, 0.5, 1, 2, 4, 8, 16, 32, 64, \infty]$ seconds. For each method separately, we divide the networks into these intervals considering the Future Window. A representative value of each interval was selected based on the highest results within each interval. Notice that a few intervals may be left empty. 

(3) In the third study, we analyzed the \emph{marginal performance} and \emph{marginal importance} of the investigated hyperparameters on the F1@50 using the fANOVA method \cite{hutter2014efficient}. The receptive field is determined by the hyperparameters of our network structure. To understand the advantages and weaknesses of each of our methods, it's essential to assess the importance and trends of the hyperparameters.

(4) In the \textit{Competitive analysis} study, we have two aims. First, it is to try to reveal what is the required delay to approach the best offline performance. Next, we aim to determine which method is more advantageous at which delay values. To this end, we need to define two new metrics : Global and local competitive ratio.
In this paper, the \emph{Competitive-Ratio}, inspired by its definition in theoretical analysis of online algorithms \cite{albers2003online}, will be the ratio between the performance of the causal (with delay) algorithm and the best acausal network. In addition, for each interval, we define the Local Competitive Ratio as the ratio between the best BF-MS-TCN++ and RR-MS-TCN++ networks in that interval.

\subsection{Results}
\begin{figure}[ht]
    \centering
    \includegraphics[width=\textwidth]{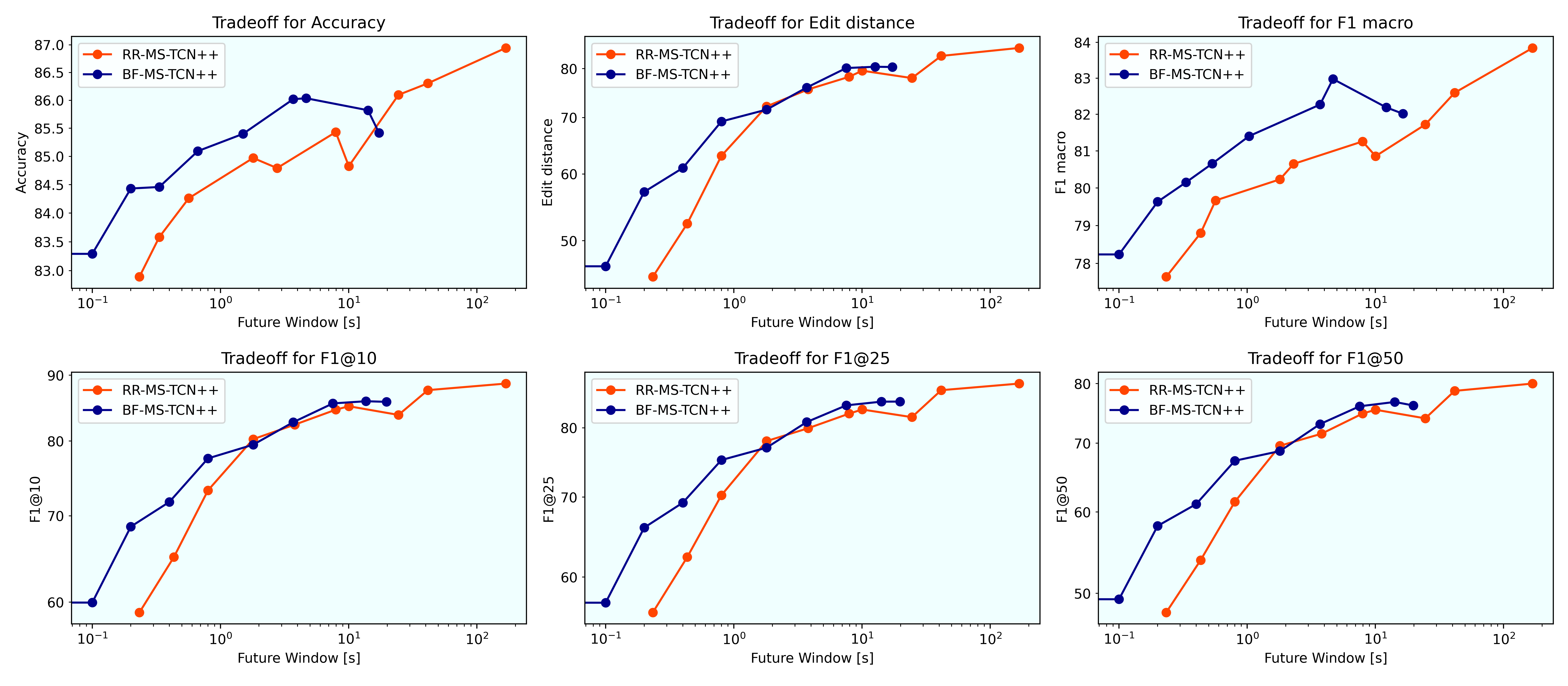}
    \caption{performance-delay trade-off. Comparison of best BF-MS-TCN++ and RR-MS-TCN++ networks, in respect to future window size. The plots show the performance in terms of Accuracy, Edit distance, F1-Macro, and $F1@k,\,\:k \in\{10,25,50\}$.}
    \label{fig:results_bestTCN}
\end{figure}

Table \ref{table:results_standard_nets} lists the results of the baseline estimation study. The feature extractor performed relatively well in a frame-wise manner with an accuracy of $82.66$ and F1-Macro of $79.46$. Both the causal and acausal networks improve the accuracy and F1-Macro scores, however, the acausal network has a significant effect in both frame-wise metrics while in the causal case only the accuracy has been improved significantly. The Causal case has the best results in all metrics. 
Fig.\ref{fig:results_bestTCN} shows a performance comparison study, where the trend is similar for all metrics, where BF-MS-TCN++ outperforms RR-MS-TCN++, especially for small delay values.

\begin{figure}[ht]
    \centering
    \includegraphics[width=\textwidth]{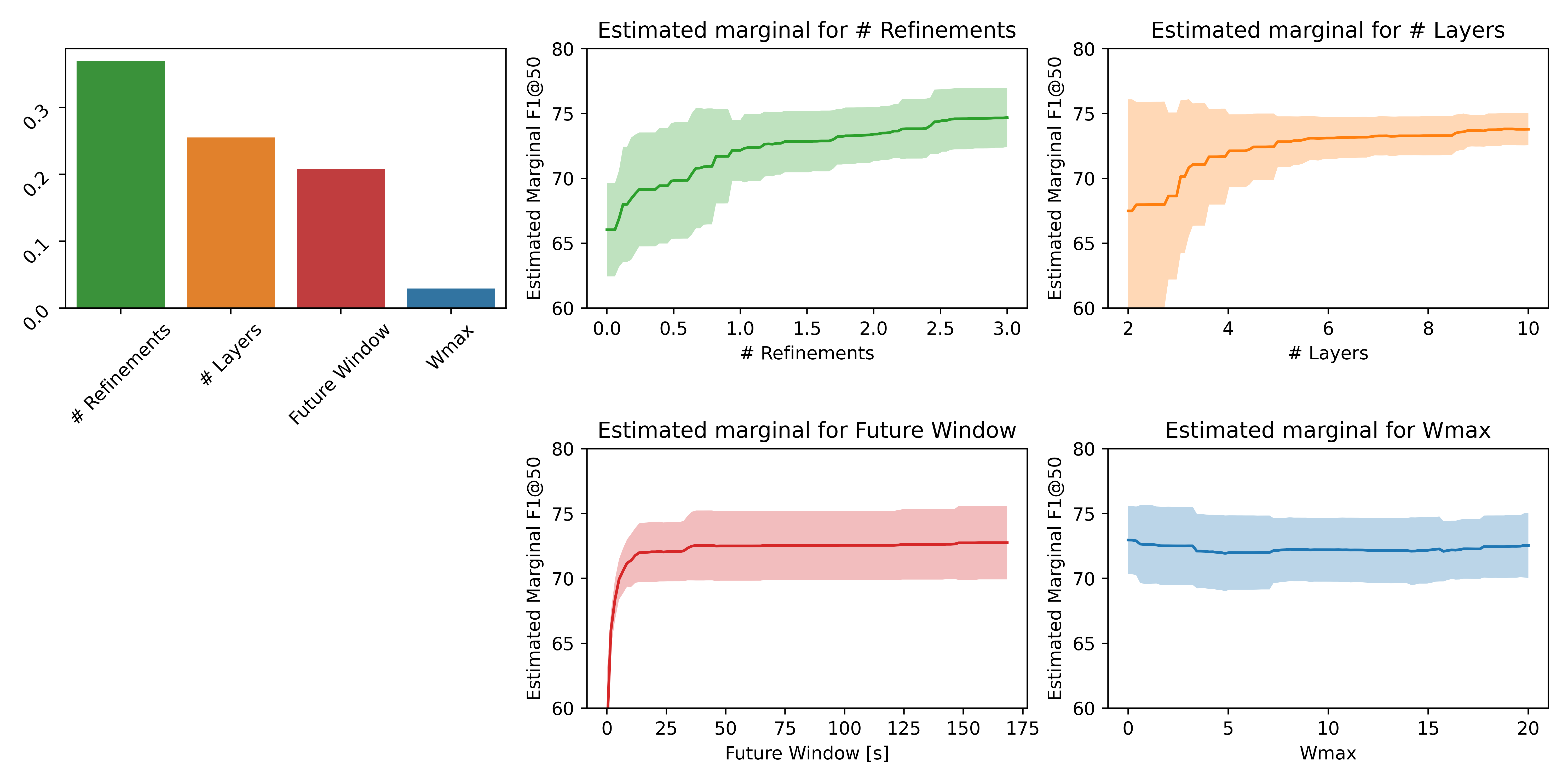}
    \caption{fANOVA analysis of BF-MS-TCN++. Marginal importance (leftmost graph) and estimated marginal F1@50 on architecture hyperparameters: number of refinement stages, number of (D)DRLs in every stage, the Future Window of the network, and the value of $w_{max}$.}
    \label{fig:fANOVA}
\end{figure}

\begin{figure}[ht]
    \centering
    \includegraphics[width=\textwidth]{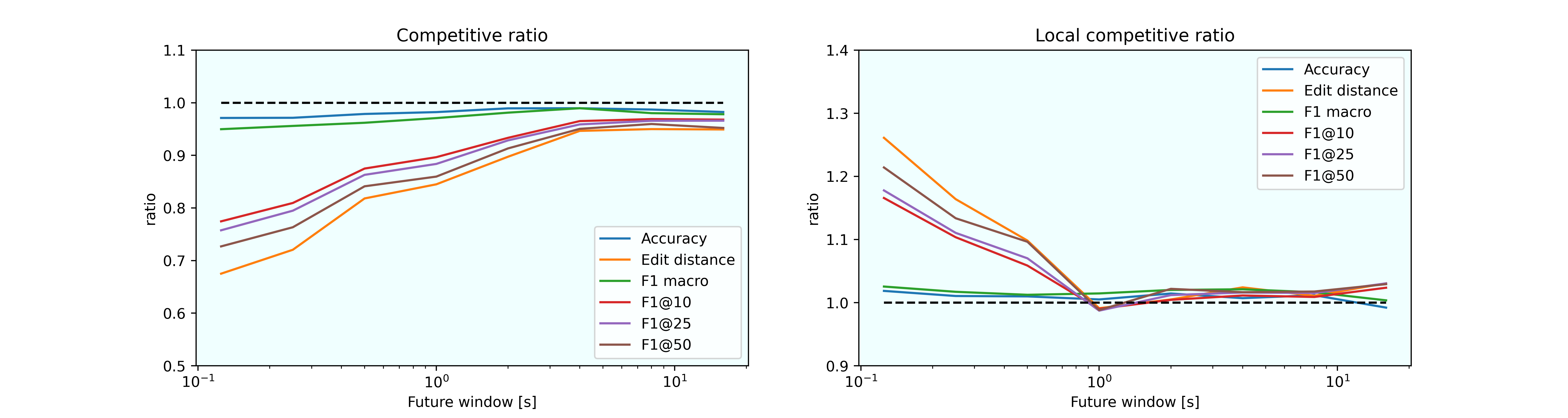}
    \caption{Competitive ratio analysis between BF-MS-TCN++ and RR-MS-TCN++, for best performing network (left) and best performing with respect to future window (right) vs time [s] (X-Axis). Competitive ratio larger than 1 means that BF-MS-TCN++ performs better than RR-MS-TCN++. The dotted black line denotes the baseline (No competitive advantage) }
    \label{fig:compratio}
\end{figure}

In the feature importance study (Fig. \ref{fig:fANOVA}), $w_{max}$ was found to have negligible importance. Other hyperparameters perform better when their values are increased.
Lastly, the Competitive analysis results are illustrated in Fig.\ref{fig:compratio}.
In the global analysis (left) plot, it is evident that BF-MS-TCN++ has 80\% of the performance of the best offline algorithm after only one second and 90\% after $2\frac{1}{3}$ seconds.
In the local analysis (right) plot, it is seen clearly that for future windows smaller than one second the BF-MS-TCN++ is significantly preferred over the RR-MS-TCN++.

\begin{table}[t]

\centering
\resizebox{\columnwidth}{!}{
\begin{tabular}{ccccccc}
\hline
\textbf{}        & \textbf{$F_1$-Macro} & \textbf{Accuracy}     & \textbf{Edit distance}     & \textbf{F1@10}    & \textbf{F1@25}    & \textbf{F1@50}    \\ \hline
EfficientNet v2  & $79.46 \pm 8.10$  & $82.66 \pm 6.03$ & -                 & -                 & -                 & -                 \\ \hline
Causal MS-TCN++  & $80.42 \pm 8.67$  & $85.04 \pm 5.77$ & $64.69 \pm 12.36$ & $74.30 \pm 10.85$ & $72.34 \pm 12.04$ & $64.96 \pm 14.07$ \\ \hline
Acausal MS-TCN++ & \textbf{83.85 $\pm$ 8.90} & \textbf{86.94 $\pm$ 6.50} & \textbf{84.65 $\pm$ 9.25}  & \textbf{88.66 $\pm$ 7.79}  & \textbf{87.13 $\pm$ 9.03}  & \textbf{80.01 $\pm$ 13.21} \\ \hline

\end{tabular} 
}
\caption{The Feature extractor EfficientNet v2, causal and acausal MS-TCN++ results on a gesture recognition task. Bold  denotes best results for metric.}
  \label{table:results_standard_nets}

\end{table}


\section{Discussion and Conclusions}
Automated workflow analysis may improve operating room efficiency and safety.
Some applications can be used offline after the procedure has been completed, while other tasks require immediate responses without delay. Nevertheless, some applications require real-time yet may allow a delayed response, assuming it improves accuracy.

 Studies showed that there is a performance gap between causal and acausal systems
 \cite{zhang2021swnet}.
To choose the optimal network that compromises between delay and performance, it is necessary to investigate how the delay affects performance.
Funke \etal  \cite{funke2019using} investigated the effect of delay on a 3D neural network. They found that adding delay improves the system's performance, primarily in segmentation metrics such as segmental edit distance and F1@10. Nevertheless, today these networks are considered relatively weak compared to the newer activity recognition networks that are typically based on transformers and temporal convolutional networks. In these algorithms, designing a future window is not trivial as in the 3D convolutional case.

In this work, we developed and analyzed different variations of the MS-TCN++, and studied the trade-off between delay and performance.
This study sought to verify the intuition that adding a relatively small delay in the causal system's response, which usually operates in real-time, could also help bridge the causal-acausal gap in MS-TCN++.
We examined this hypothesis in two methods. First, we tried to examine reducing the network's depth, thus the receptive field was reduced, with half of it being the future, namely Reduced Receptive field MS-TCN++, RR-MS-TCN++. 
We expected that this method would not work well for small future window sizes because reducing the window size in this method means a significant decrease in the number of layers or even eliminating of few refinement stages. Li \etal \cite{li2020mstcnpp}, showed these parameters have a crucial effect on performance.
Therefore we developed the BF-MS-TCN++. This network involved applying a convolution in which the target point in the previous layer is not in the middle of the convolution's receptive field. This way, we bound the future window of the entire network even in large architectures with a small delay. With this method, different architectures can be implemented for approximately the same delay values.
Thus in our analysis, several time intervals were defined for both methods, and we selected the network with the best results to represent each interval.

According to Fig.\ref{fig:results_bestTCN}, the performance comparison study illustrates the trade-off between performance and delay, considering all metrics in both methods. As seen in Table \ref{table:results_standard_nets}, the gap between causal and acausal networks is much more prominent in the segmentation metrics than in the frame-wise metrics. Similarly, in Fig. \ref{fig:results_bestTCN} and Fig. \ref{fig:compratio}, these metrics exhibit a stronger trade-off effect. Furthermore, BF-MS-TCN++ outperforms RR-MS-TCN++ especially when small delays are allowed.
In the left image of Fig.\ref{fig:compratio}, a future window of $2\frac{1}{3}$ seconds achieves more than 90\% of the best offline network. Namely, getting close to an offline network's performance is possible even with a relatively small delay.

The fANOVA test showed, in Fig.\ref{fig:fANOVA}, that the number of (D)DRLs and the number of refinement stages are the most important hyperparameters for optimizing performance, even more than the total delay. Where increasing these hyperparameter values tends to improve estimated marginal performance. Thus our BF-MS-TCN++, which allows the design of larger networks with that same delay factor, takes advantage of this fact. Another interesting finding is that the value of $w_{max}$ barely affects the outcome (Fig. \ref{fig:fANOVA}), whereas the total future delay plays a pivotal role. Thus, we conclude that minimizing the value of $w_{max}$ for designing larger networks is an acceptable approach.

This study has a few limitations. First, it has been evaluated using only one data set. In addition, only data from surgical simulators were analyzed. Larger data sets, including data from the operating room, should be analyzed in the future. 

The algorithms presented in this study are not limited to the surgical domain. Online with delay activity recognition is relevant to many other applications that evaluate human performance, even outside the surgical field. Therefore, this study lays the foundations for a broader study on the relationship between accuracy and delayed response in video-based human activity recognition.

%
%
\bibliographystyle{splncs04}
\bibliography{egbib}
\end{document}